\documentclass{article}
\usepackage{ijcai20}

\usepackage{times}

\usepackage{soul}
\usepackage{url}
\usepackage[hidelinks]{hyperref}
\usepackage[utf8]{inputenc}
\usepackage[small]{caption}
\usepackage{graphicx}
\usepackage{amsmath}
\usepackage{booktabs}
\title{Label-guided Learning for Text Classification}

\begin{document}

\maketitle

%
\begin{abstract}
Text classification is one of the most important and fundamental tasks in natural language processing.   Performance of this task mainly dependents on text representation learning. Currently, most existing learning frameworks mainly focus on encoding local contextual information between words.  These methods always neglect to exploit global clues, such as label information,  for encoding text information. In this study, we propose a label-guided learning framework {\bf LguidedLearn} for text representation and classification. Our method is novel but simple that we only insert a label-guided encoding layer into the commonly used text representation learning schemas.  That label-guided layer performs label-based attentive encoding to map the universal text embedding (encoded by a contextual information learner) into different label spaces, resulting in label-wise embeddings.  In our proposed framework, the label-guided layer can be easily and directly applied with a contextual encoding method to perform jointly learning.  Text information is encoded based on both the local contextual information and the global label clues. Therefore, the obtained text embeddings are more robust and discriminative for text classification.  Extensive experiments are conducted on benchmark datasets to illustrate the effectiveness of our proposed method.
\end{abstract}

\section{Introduction}

Text classification  can be simply described as the task that given a sequence of text (usually a sentence, paragraph, or document\footnote{We will not specifically differentiate ``sentence", ``document", and ``paragraph". These terms can be used interchangeably.}) we need to build a learning system to output a one-hot vector\footnote{In this study we  only consider single label (not multi-labels) classification problem.} to indicate the category/class of the input sequence of text. It is a very important and fundamental task in the natural language processing community. 
In practice,  many real applications can be cast into  text classification tasks, such as document organization, news topic categorization, sentiment classification, and text-based disease diagnoses \cite{liu2016recurrent,miotto2016deep,wang2018joint,pappas2019gile,yao2019graph}.

The essential step in text classification is to obtain text representation.  In the earlier study, a piece of given text is usually represented with a  hand-crafted feature vector \cite{wang2012baselines}. 
Recently, inspired by the success of word embedding learning, a piece of text (a sentence/paragraph/document) is also represented with an embedding, which is automatically learned from the raw text by neural networks. Theses learning methods mainly include sequential-based learning models \cite{kim2014convolutional,zhang2015character,conneau2016very,liu2016recurrent} and graph-based learning models \cite{kipf2016semi,cai2018comprehensive,battaglia2018relational,yao2019graph}.  All of these text learning methods are based on modeling local contextual information between words to encode a piece of text into a universal embedding, without considering the difference of labels.
    More recently, some research suggests that global label clues are also important for text representation   learning \cite{rodriguez2013label,akata2013label,nam2016all,wang2018joint,pappas2019gile}.

  In this study, we exploit label constraints/clues to guide text information encoding and propose  
  a label-guided learning framework {\bf LguidedLearn} for text classification. In our framework, each label is represented by an embedding matrix. A label-guided layer is proposed to map universal contextual-based embeddings into different label spaces, resulting in label-wise text embeddings. LguidedLearn performs jointly learning of word-word contextual encoding and label-word attentive encoding. The ultimately obtained text embeddings are informative and discriminative for text classification. A series of comprehensive experiments are conducted to illustrate the effectiveness of our proposed model.

\section{Related Work}
In the computer vision community, many studies have exploited label information for image classification \cite{akata2013label,frome2013devise}. All of these models jointly encode label description text information and image information to enhance the performance of image classification. Recently, several studies have involved label embedding learning in natural language processing tasks. For example, Nam \emph{et al.} \shortcite{nam2016all} proposed a model to learn the label and word embeddings jointly. Pappas \emph{et al.} \shortcite{pappas2019gile} also presented a model GILE to encode input-label embedding. However, all these models require that there must be a piece of description text for each label. The learning performance is dependent on the quality of the label description text. Furthermore, this requirement will limit the models' application.
%

%
%
\section{Method}
In this section,  we first intuitively describe how to involve labels into text encoding and the main layers needed for text representation learning. Then, we present formally the proposed framework LguidedLearn (Label-guided Learning) for text classification.

\begin{figure}[h!bt]
	\centering
	\includegraphics[scale=0.6]{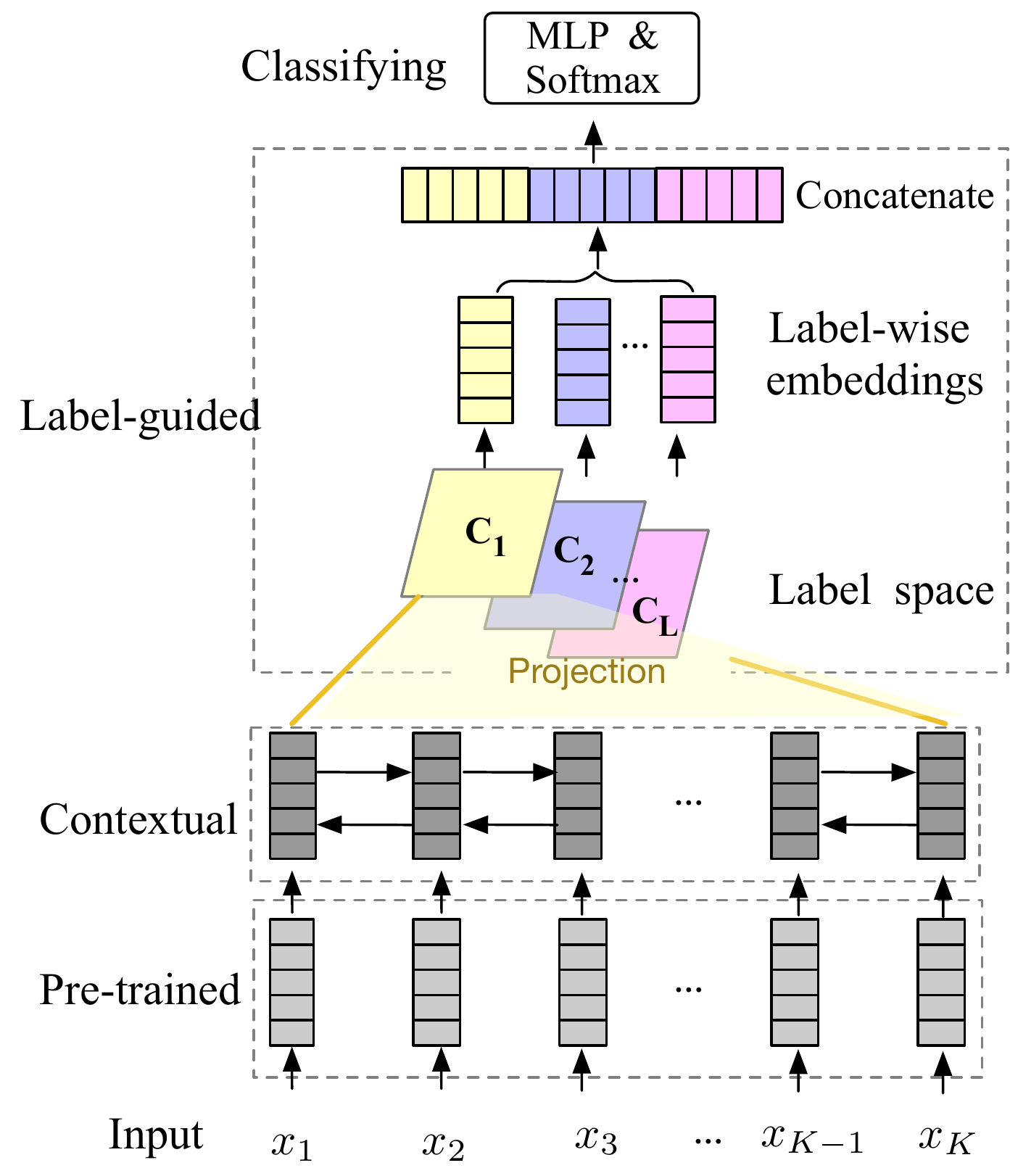}
	\caption{The framework of our proposed label-guided learning, {\bf LguidedLearn}, for text classification.}
	\label{fig:framework}
\end{figure}

\subsection{Intuition}
Given a piece of text, what kinds of information/clues should be encoded in the text representation learning for classification? First local contextual information is essential for text embedding learning. We notice that not all words/characters in the given text are equally useful for correctly labeling. Therefore, we also need global label constraints/information to guide text encoding.   Based on the above considerations, a learning framework for text classification should include:
\begin{itemize}
	\item[-] Pre-trained encoding layer:  get pre-trained word or character embeddings;
	\item[-] Contextual encoding layer:  encode contextual information between words  into the text embeddings
	\item[-] Label-guided encoding layer: perform label attentive learning to encode global information (constraints) into the text embeddings; 
	\item[-] Classifying layer: conduct feature compression and text classification. 
\end{itemize}

\subsection{The Framework: LguidedLearn}
 The proposed label-guided learning framework is shown in the Figure \ref{fig:framework}. In this section, we introduce the framework    in detail. 
 Given a sequence of text ($ x_1, x_2, \cdots, x_K $), we apply the following learning layers successively.
 \subsubsection{Pre-trained encoding layer} The aim of pre-trained layer is to obtain low-dimensional continuous embeddings for words in the sequence of text. 
 \begin{equation}
 	 (e_{x_1}, e_{x_2}, \cdots, e_{x_K}) = f_{\text{pre-trained}}(x_1,  x_2, \cdots, x_K ),
 \end{equation}
 where $e_{x_i}\in R^{m_p\times 1}$ (where $m_p$ is the pre-trained embedding size)  is a pre-trained embedding of word $x_i$, and  $ f_{\text{pre-trained}}$ is a kind of word embedding learner, such as  Glove \cite{pennington2014glove}. 

 \subsubsection{Contextual encoding layer} The contextual layer further encodes words' dynamic contextual information in the current text sequence.
 \begin{equation}
 	   	 (\bar e_{x_1}, \bar e_{x_2}, \cdots, \bar e_{x_K}) = f_{\text{contextual}}   	 (e_{x_1}, e_{x_2}, \cdots, e_{x_K}),  
\end{equation}
  where $\bar e_{x_i}\in R^{m_c\times 1}$ (where $m_c$ is the contextual embedding size) is a contextual-encoded embedding corresponding to the  word $x_i$. In the supervised learning tasks,  $f_{\text{contextual}}$ can be effectively implemented with a LSTM or BiLSTM network. 

 \subsubsection{Label-guided encoding layer}
 
 Let $\{l_1, l_2, \cdots, l_L  \}$ be the label set, where $L$ is the number of labels. Each label $l_i$ ($i=1,2, \cdots, L$) is represented with an embedding matrix $C_{l_i} \in R^{m_l\times t}$ consisting of $t$ embeddings ($c_{l_i,1}, c_{l_i,2},\cdots, c_{l_i,t}$), where $c_{l_i,j} \in R^{m_l\times 1}$  ($j=1,2,\cdots, t$) is the $j$th embedding in the embedding matrix $C_{l_i}$. The label-guided layer jointly encodes label information and contextual information by projecting contextual-encoded embeddings into the  label space. 
 Take the $i$th ($i=1,2, \cdots, L$) label for example:
  \begin{equation}
 	 v_{{l_i}} = f_{\text{Lguided}} (\bar E_{\text{x}}, C_{l_i}), \  \  \ 
 \end{equation}
 where 
   $\bar E_{\text{x}} =  (\bar e_{x_1}, \bar e_{x_2}, \cdots, \bar e_{x_K}) $  are the contextual-encoded embeddings of the given sequence text, $v_{{l_i}}$ is a label-wise embedding specified with the $i$th label.  $f_{\text{Lguided}}$ can be implemented with the following simple way:  
    \begin{equation}
  v_{l_i} =  \sum_{j=1}^K \bar w_{l_i,j} \bar e_{x_j},
  \label{equ:label-wise-emb}
   \end{equation}
  where $\bar w_{l_i,j} $ is a label attentive weight to measure the compatibility of the pair $<\bar e_{x_j}, C_{l_i}>$, where $\bar e_{x_j}$ is the 
  contextual embedding of the $j$th word in the given sequence text, 
  and $C_{l_i}$ is the embedding matrix of the label $l_i$. To get the compatibility weight,  the cosine similarities between  $\bar e_{x_j}$ and each embedding in $C_{l_i}$ = ($c_{l_i,1}, c_{l_i,2},\cdots, c_{l_i,t}$) are computed \footnote{To make the cosine similarities can be computed, the dimension of contextual embeddings must be equal to the dimension of the embeddings in the label embedding matrix. That is $m_c= m_l$.}, resulting in a similarity degree vector ($s_1, s_2,\cdots, s_t$). The largest similarity value is collected by 
  \begin{equation}
  	w_{l_i,j} = \text {max-pooling} (s_1, s_2,\cdots, s_t), \    \    \  j=1,2, \cdots, K
  \end{equation}
 Considering the formula (\ref{equ:label-wise-emb}), the collected values ($w_{l_i,1}, w_{l_i,2}, \cdots, w_{l_i,K}$) should be normalized 
 \begin{equation}
 	(\bar w_{l_i,1}, \bar w_{l_i,2}, \cdots, \bar w_{l_i,K}) = \text {SoftMax} (w_{l_i,1}, w_{l_i,2}, \cdots, w_{l_i,K})
 \end{equation}
 
According to the label-guided encoding formula (\ref{equ:label-wise-emb}), we can obtain  label-wise embeddings ($v_{l_1}, v_{l_2}, \cdots, v_{l_L}$), where  $v_{l_i}$ ($i=1,2, \cdots, L$) is the text embedding encoded with the guidance of the label $l_i$.  All the label-wise embeddings are concatenated 
  \begin{equation}
  	v = \text{concate}( v_{{l_1}}, v_{{l_2}}, \cdots, v_{{l_L}} ),
  \end{equation}
where $v$ is the ultimate embedding used to represent the given sequence of text.

\subsubsection{Classifying layer} In the classifying layer, we first employ $\text{MLP}$ to compress the large text embedding $v$ into a proper one
\begin{equation}
\bar v = \text{ReLU} (\text{MLP}(v)),
\end{equation}
where $\bar v \in R^{m_f \times 1}$. In this study, we empirically set $m_f = 10 L$ (10 times of the number of the labels). Then the classifying label distribution  of the given sequence of text is obtained by
\begin{equation}
\bar y = \text{SoftMax} (\text{MLP}(\bar v))
\end{equation}
In the above formula, $\text{MLP}$ is used to encode the embedding $\bar v$ into a vector with dimension equal to the number of the labels.

\section{Experiments and results}

\subsection{Datasets}
In this study, we employ two text datasets to investigate the effectiveness of our proposed  method. For convenience,  the two  datasets are simply denoted by {\bf TCD-1} (TCD: Text Classification Dataset) and {\bf TCD-2}, respectively. A summary statistics of the two datasets (TCD-1 and TCD-2) are presented in Table \ref{tab:test-sets}.  
%
\begin{table}[tbh] 
\small
\centering
\caption{Summary statistics of the  datasets TCD-1 and TCD-2. Note: \#Doc is the number of documents and \#Len is the average length of each document.}
	\begin{tabular}{@{}c|ccccc@{}}
    \hline
	\bf{Dataset} &  \bf{Name} & \bf{\# Doc} 	 &  \bf{\# Word} &  \bf{\# Class}   &  \bf{\# Len}  \\
	\hline
& DBPedia&  630,343&   21,666&  14&  57.31   \\
& Yelp-B &  598,000&   25,709& 2&  137.97   \\
TCD-1 & Yelp-F &  700,000&    22,768&  5&  151.83  \\
& YahooQA&   1,460,000&      607,519&   10&  53.40   \\
& AGNews &  127,600&    13,009&  4&  43.84 \\
\hline
	& 20NG  &18,846&    42,757&20&221.26  \\
& R8&  7,674    &7,688   &8   & 65.72   \\
TCD-2  & R52   &9,100    &8,892  &52  &69.82  \\
& Ohsumed&   7,400&    14,157&  23&  135.82  \\
& MR&   10,662   &18,764  &2  & 20.39  \\
	\hline
	\end{tabular}
\label{tab:test-sets}
\end{table}
 \paragraph{TCD-1}
 We select TCD-1 as an experimental dataset  because it is used by \cite{wang2018joint} to 
     evaluate their proposed  text and label jointly learning model LEAM, which is a main baseline for comparative analysis. TCD-1 includes five sub-datasets:     
\begin{itemize}
	\item[-] {\bf AGNews}: news topic classification over four categories (world, entertainment, sports and business).  Each document is composed of an internet news article.
	\item[-] {\bf Yelp-F}: sentiment classification of polarity star labels ranging from 1 to 5. The dataset is obtained from the Yelp Review Dataset Challenge in 2015. 
	\item[-] {\bf Yelp-B}: sentiment classification of polarity labels (negative or positive). The dataset is the same as the Yelp Full, only the labels are different, where polarity star 1 and 2 are treated as negative, and 4 and 5 as positive.
	\item[-] {\bf DBPedia}: ontology classification over fourteen classes. The dataset is picked from DBpedia 2014 (Wikipedia).
	\item[-] {\bf YahooQA}: QA topic classification over ten categories. The dataset is collected from the version 1.0 of the Yahoo! Answers Comprehensive Questions and Answers.
     \end{itemize}

\begin{figure}[h!tb]
	\centering
	\includegraphics[scale=0.5]{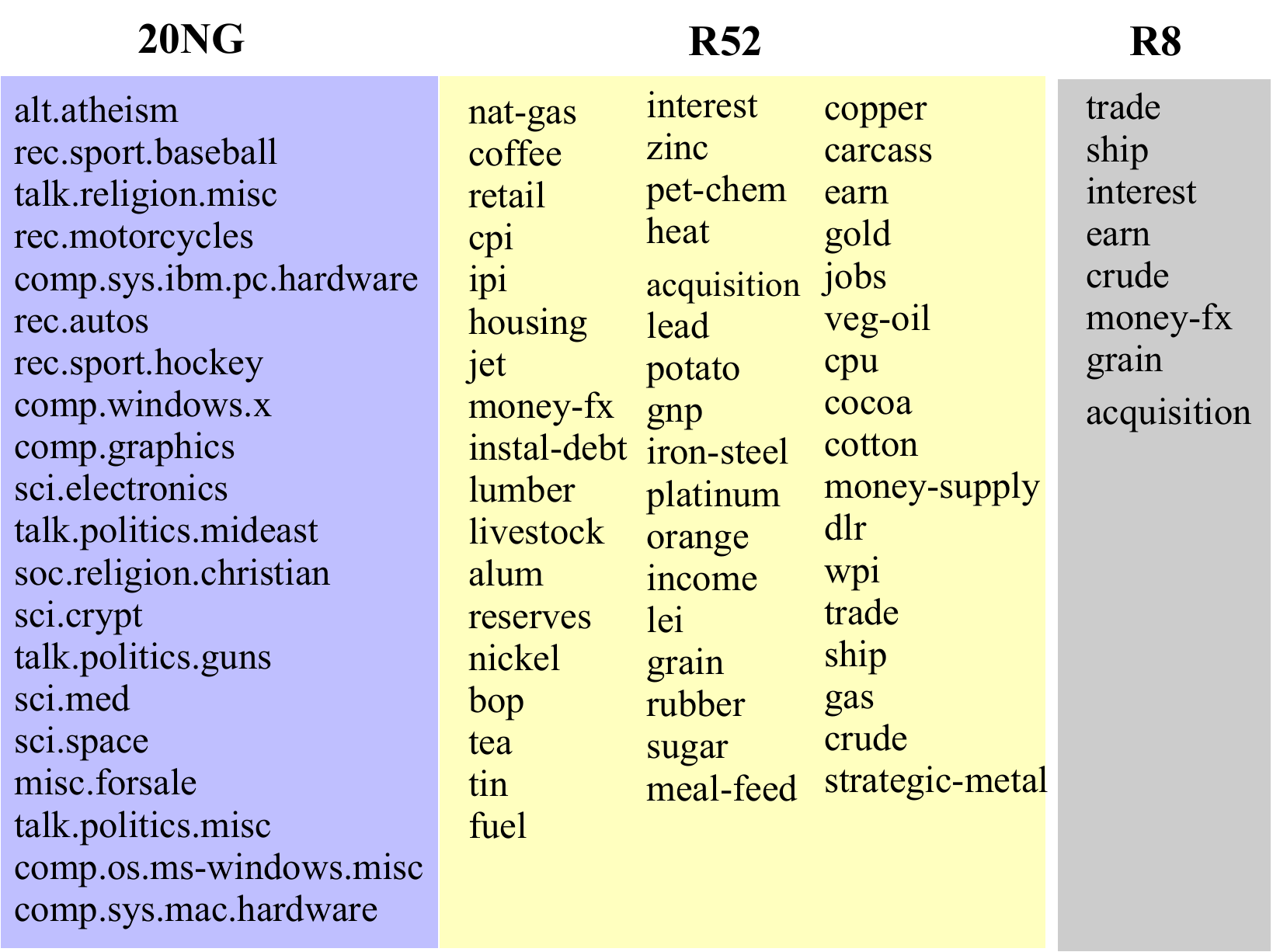}
	\caption{Labels of the document topic classification datasets: 20NG, R52 and R8.}
	\label{fig:tcd-2-label}
\end{figure}

 \paragraph{TCD-2}
We select TCD-2 to test our model because it is a very comprehensive dataset, referring to many different domains, and is widely used by more recent work \cite{yao2019graph}. TCD-2   also consists of five sub-datasets:
\begin{itemize}
	\item[-] {\bf 20NG}: document topic classification over 20 different categories.
    \item[-] {\bf R52}: document topic classification over 52 categories. The R52 dataset is a subset of the Reuters 21578 dataset. 
	\item[-] {\bf R8}: document topic classification over 8 categories. The R8 dataset is also a subset of the Reuters 21578 dataset. 
	\item[-] {\bf Ohsumed}: disease classification over 23 categories. The Ohsumed corpus is from the MEDLINE database.
	\item[-] {\bf MR}: binary sentiment classification. The MR dataset is a movie review dataset, in which each review only contains one sentence.
\end{itemize}
Though 20NG, R52 and R8 are all document topic classification. The three sub-datasets are very different in labels, as presented in Figure \ref{fig:tcd-2-label}. For example, 20NG mainly refers to sports, politics and computers. While R52 and R8 mainly contains life related issues and financial related topics. R8 has a more rough categorizations than that of R52.  
%
%

\subsection{Baselines}
\paragraph{Baselines on TCD-1}
Baselines reported on TCD-1 mainly contain five categories:
\begin{itemize}
	\item[-] Powerful traditional language feature models, such as BOW (bag-of-words) \cite{zhang2015character};
	\item[-] Effective embedding based models, including fastText \cite{joulin2016bag} and SWEM \cite{shen2018baseline};
	\item[-] Deep learning models, mainly including commonly used CNN-based models and LSTM-based models, such as Small CNN \cite{zhang2015character}, Large CNN \cite{zhang2015character}, Deep CNN \cite{conneau2016very}, LSTM, and SA-LSTM;
	\item[-] Attention based models: HAN \cite{yang2016hierarchical} and Bi-BloSAN \cite{shen2018bi};
	\item[-] Label embedding based models: LEAM \cite{wang2018joint}  and our proposed LguidedLearn. 
\end{itemize}

\paragraph{Baselines on TCD-2}
Baselines conducted on TCD-2 are:
\begin{itemize}
	\item[-] Traditional models: TF-IDF + LR;
	\item[-] Embedding based models: PV-DBOW \cite{le2014distributed}, PV-DM  \cite{le2014distributed}, fastText \cite{joulin2016bag}, and SWEM \cite{shen2018baseline};
	\item[-] Sequential deep learning models: CNN-rand, CNN-non-static and  LSTM;
	\item[-] Graph deep learning models: Graph-CNN-C \cite{defferrard2016convolutional}, Graph-CNN-S \cite{bruna2013spectral}, Graph-CNN-F \cite{henaff2015deep} and Text GCN \cite{yao2019graph};
	\item[-] Attention based models, HAN \cite{yang2016hierarchical} and Bi-BloSAN \cite{shen2018bi};
	\item[-] Label embedding based model LEAM \cite{wang2018joint} and our proposed LguidedLearn. 
\end{itemize}
TCD-2 is a very comprehensive dataset used for text classification.  Besides sequential deep learning models, some recent graph-based neural network models 
have also been reported on TCD-2.

\subsection{Experimental settings}
\paragraph{Dataset Setting}
The settings of the training/testing of TCD-1 and TCD-2 are as the same of the used in \cite{wang2018joint} and \cite{yao2019graph}, respectively. 
\paragraph{Model setting}
In the pre-trained layer, we use Glove to obtain pre-trained word embeddings with dimension of  $m_p=300$. The contextual layer is implemented by BiLSTM with dimension of $m_c=300$. In the label-guided layer, each label is represented by an embedding matrix with size of $m_l\times t = 300\times 5$. That is, each label is represented by five embeddings. 
\paragraph{Learning setting} In our training,  batch size is 25 and  learning rate is 0.001. All experiments are repeated 10 times.

\subsection{Results and analysis}

\subsubsection{Performance comparison}
A series of comparison experiments are conducted on the datasets of TCD-1 and TCD-2. A summary of text classification performance and simple analysis  are presented. 

\paragraph{- Results on TCD-1}
The comparison results on TCD-1 are presented in Table \ref{tab:compare-TCD-1}. The  results illustrate that our proposed framework LguidedLearn can achieve the best performance  on  all the test datasets of TCD-1. 
Even compared to some recently published strong text classification algorithms (such as fastText, SWEM,  Deep CNN (29 layers), Bi-BloSAN, and LEAM) LguidedLearn can obtain stable and prominent gains, especially on the AGNews, Yelp-B, and  Yelp-F datasets. 
 
%
%
\begin{table}[h!tb]
\centering
\small
\caption{The results of text classification accuracy (\%) on TCD-1.}
	\begin{tabular}{@{}c|c@{}c@{}c@{}c@{}c@{}}
    \hline
{ Model} \, & { YahooQA} \, & { DBPedia	} \, & { AGNews	} \,  & { Yelp-B} \, & {\ Yelp-F} 	\\
\hline
	BOW  & 68.90 & 96.60 &88.80 & 92.20 & 58.00 \\
\hline
	SWEM   & 73.53 & 98.42 & 92.24 &93.76 & 61.11 \\
	fastText   & 72.30 & 98.60 & 92.50  & 95.70 & 63.90 \\
\hline
	Small CNN & 69.98 & 98.15  &89.13 & 94.46 &58.59 \\
	Large CNN   & 70.94 &98.28 &91.45 & 95.11 & 59.48  \\
	Deep CNN    & 73.43 &98.71 & 91.27 &95.72 & 64.26 \\
	LSTM  & 70.84 & 98.55 & 86.06 &94.74 &58.17 \\
	SA-LSTM    &- & 98.60 &- &- & - \\
\hline
	HAN  & 75.80 & - &- &- &- \\
	Bi-BloSAN  & 76.28 & 98.77 & 93.32 & 94.56 & 62.13 \\
\hline
	LEAM-linear & 75.22 & 98.32 & 91.75 & 93.43 & 61.03 \\
	LEAM  & 77.42 & 99.02& 92.45& 95.31 &64.09 \\
	LguidedLearn & {\bf 77.61} & {\bf 99.08} 	& {\bf 93.67} &	{\bf 96.80}	& {\bf 68.08 }\\
	\hline
	\end{tabular}
\label{tab:compare-TCD-1}
\end{table}

\paragraph{- Results on TCD-2}
 The results  conducted on TCD-2 are presented in Table \ref{tab:compare-TCD-2}. From the results we can see that besides the dataset of 20NG, LguidedLearn obtains the best classification accuracy on all the other datasets. The recent very popular graph neural network (GNN) models show strong ability in classifying text documents. We notice that even the very strong GNN-based models (such as Graph-CNN-C, Graph-CNN-S, Graph-CNN-F, and Text GCN) are surpassed by our proposed framework LguidedLearn by a large margin. Documents of 20NG are very long (where average length is  221.26 words and about 18\% documents are more than 400 words) and surpass the  maximum encoding length  of the majority models.  This is the reason that on the dataset of 20NG  non-sequential encoding models (TF-IDF+LR, SWEM, and Text GCN) can achieve better performance than those sequential encoding models (such as LSTM and CNN based models). LguidedLearn (having a contextual layer with BiLSTM) is also slightly affected by this factor. 
\begin{table}[h!tb]
\centering
\caption{The results of text classification accuracy (\%) on TCD-2.}
	\begin{tabular}{@{}c@{}|c@{}c@{}c@{}c@{}c@{}}
    \hline
	{ Model } & {20NG}  &  {R8} \, &  {R52}  \, 	& {Ohsumed}  &	{MR}	 \\
\hline
TF-IDF + LR &	\,  \,  \,  83.19   \,  \,  \, &  93.74   \,  \, & 86.95  \,  & 54.66   \,  & 74.59  \\
\hline
PV-DBOW	& 74.36   &  85.87  &  78.29   &  46.65   & 61.09  \\
PV-DM&	51.14  &  52.07  & 44.92   & 29.50   & 59.47  \\
fastText &	11.38   & 86.04  & 71.55  & 14.59  & 72.17 \\
SWEM &	85.16  & 95.32  & 92.94  & 63.12  & 76.65  \\
\hline
CNN-rand& 76.93 &  94.02 &  85.37 &  43.87  &  74.98  \\
CNN-non-static&	82.15   &  95.71   &  87.59   &  58.44  &  77.75 \\
LSTM&	65.71  &  93.68  &  0.8554   &  0.4113   &  0.7506 \\
LSTM (pretrain)	& 75.43  &  96.09  & 90.48  & 51.10   & 77.33  \\
\hline
Graph-CNN-C &	81.42  &  96.99  & 92.75 & 63.86  & 77.22  \\
Graph-CNN-S &	$-$  & 96.80  &  92.74  & 62.82  & 76.99  \\
Graph-CNN-F&	$-$  & 96.89  & 93.20  & 63.04  & 76.74  \\
Text GCN &	\bf {86.34 } &   97.07  & 93.56  & 68.36  & 76.74 \\
\hline
 LEAM &	81.91  & 93.31  & 91.84  & 58.58  & 76.95   \\
 LguidedLearn &  85.58   & {\bf 97.86 }  &  \bf{96.12 }  &  \bf{70.45 } &  \bf{82.07 } \\
	\hline
	\end{tabular}
\label{tab:compare-TCD-2}
\end{table}

\subsubsection{Analysis of Label-guided encoding}
The main difference between our presented LguidedLearn and the traditional deep learning models is  the label-guided encoding layer, which performs a label attentive learning.   Compared to previous label attentive learning models, such as LEAM, LguidedLearn 1) extends label embedding to label embedding space (represented by a series of embeddings) and 2) performs jointly learning of contextual encoding and label-guided encoding.   We present a series of comprehensive analysis to the label-guided layer according to the above considerations.

\paragraph{- Label attentive learning analysis}
The results of LguidedLearn without label-guided encoding layer,  denoted by \emph{Label-gudied (w/o)},  are presented in Table \ref{tab:analysis-ablation}. Actually, after removing label-guided layer LguidedLearn is reduced to  a BiLSTM-based text classification model. The comparison results (Label-guided (w/o) vs. LguidedLearn) show that the label-guided layer brings huge gains ($> 5\%$ accuracy improvement), especially on complex task datasets: 20NG, R52, Ohsumed and MR. 

\begin{figure}[bth]
	\centering
	\includegraphics[scale=0.6]{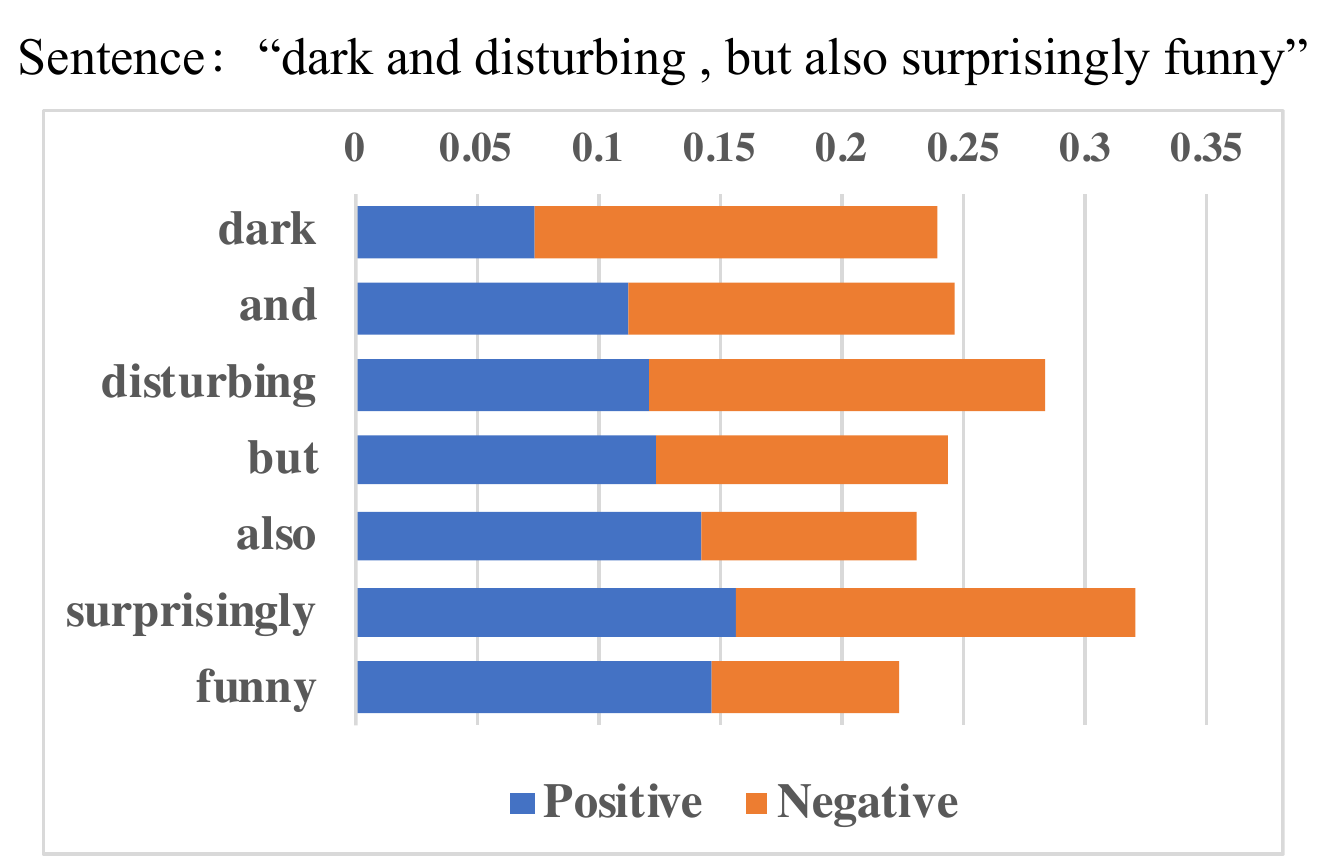}
	\caption{Visualization of the learned attentive weights between words and labels.  The sample is taken from the MR dataset which has two kinds of label: ``Positive" and ``Negative" .}
	\label{fig:case-mr}
\end{figure}

The MR dataset is a sentiment classification task which requires model having ability to capture sentiment-related detailed information. An example taken from MR is presented in Figure \ref{fig:case-mr}. The example shows that the label-guided layer can extract label (sentiment) related information from input words by performing label attentive learning. For example, words of ``funny" and ``also" are  likely to be projected into the {\bf Positive} space with large attentive weights, and words of ``dark" and  ``disturbing" are more likely projected into the {\bf Negative} space. 

\begin{figure}[h!bt]
	\centering
	\includegraphics[scale=0.7]{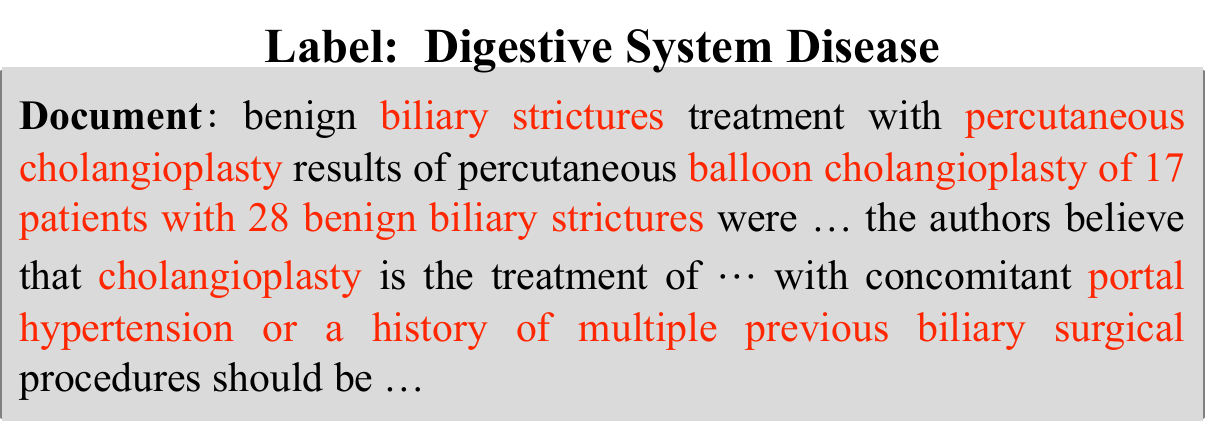}
    \caption{An example of medical text document taken from the Ohsumed dataset which has  23 different labels. The correct label of the  example is {\bf Digestive System Disease} . Words denoted by red color have large attentive weights corresponding to the correct label. The sample actually has more than 200 words (words with small attentive weights are omitted by ``...").  
     }
	\label{fig:case-ohsume}
\end{figure}

The datasets of 20NG, R52 and  Ohsumed have many labels and documents are very long. These are very challenging text classification tasks, which need model to correctly extract label-related information from a very redundant and even noisy input.  A medical text example taken from the Ohsumed dataset is presented in Figure \ref{fig:case-ohsume}. The example illustrates the effectiveness of the label attentive learning. The medical text example has  more than 200 words, only some pieces of text (denoted by red color) are effectively projected into the correct label space with large label attentive weights. These pieces of text are strongly related to the sample label {\bf Digestive System Disease}, such as ``\emph{percutaneous cholangioplasty}" and  ``{\emph balloon cholangioplasty of 17 patients with 28 benign biliary strictures}".

 
%
%
\begin{table}[h!tb]
\centering
\caption{Ablation analysis of the LguidedLearn framework. Label-guided (w/o): LguidedLearn without label-guided encoding layer.}
	\begin{tabular}{@{}c@{}|c@{}c@{}c@{}c@{}c@{}}
    \hline
	{ Model } & \, {20NG}  \, &  {R8} \, &  {R52}  \, 	& {Ohsumed}  &	{MR}	 \\
\hline 
Label-guided (w/o) &	\, 73.18  \,   \, & 96.31  \,  \,  &  90.54  \,  &  49.27  \, &  77.68 \\
LguidedLearn &  {\bf 85.58}  & {\bf 97.86 }  &  \bf{96.12 }  &  \bf{70.45 } &  \,  \bf{82.07 } \\
\hline
\end{tabular}
\label{tab:analysis-ablation}
\end{table}

%

%

\begin{figure}[bth]
	\centering
	\includegraphics[scale=0.18]{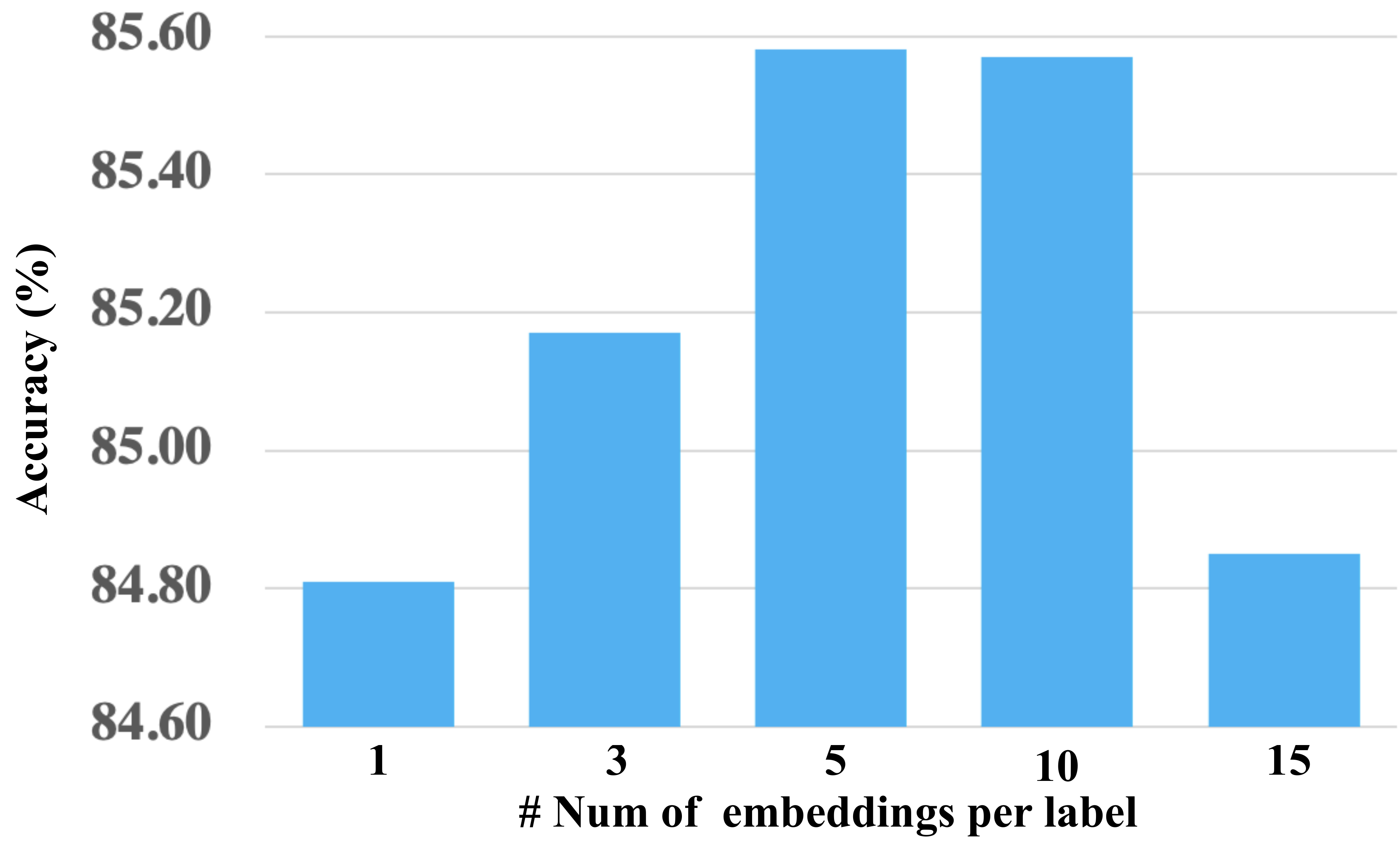}
	\caption{Results (on the 20NG dataset) of LguidedLearn with using   varying number of label embeddings.}
	\label{fig:size-label-emb}
\end{figure}

\paragraph{ - Label embedding space  analysis} 
Different from document/sentence samples, each label is actually a class, which should contains all kinds of representative information from the samples belonging to the label. 
Therefore, it's not reasonable to use only one embedding to represent a label. In the study, we extend label embedding to label embedding space which is represented by a series of embeddings (embedding matrix).  An example experiment, conducted on the 20NG dataset, is presented in Figure \ref{fig:size-label-emb} to show the accuracy performance with using varying size of label embedding matrix. From the results we can see that the performance first improves as increasing the number of embeddings per label from one to five, and then decreases with using more than ten embeddings per label.  Because at first increasing the label embedding size will increase labels` representation ability; but using too many embeddings also will decrease the label discriminative ability. The best choice of the size of label embedding matrix is dependent on datasets. For convenience and avoiding excessively fine tuning parameters, in all experiments we simply use five embeddings to represent each label (see the section of {\bf Experimental settings}). The results of a comprehensive experiment are presented in Table \ref{tab:analysis-size-label-emb}. The comparing results illustrate that using embedding matrix (used by LguideLearn) can obtain stable and fruitful improvements, compared to using one label embedding (usually used by previous label embedding model, such as LEAM). 

\begin{table}[h!tb]
\centering
\caption{Comparison results of one label embedding and label embedding space (represented by five embeddings by default in this study). LguidedLearn-1: LguidedLearn with using only one embedding to represent each label.}
	\begin{tabular}{@{}c@{}|c@{}c@{}c@{}c@{}c@{}}
    \hline
	{ Model } & \, {20NG}  \, &  {R8} \, &  {R52}  \, 	& {Ohsumed}  &	{MR}	 \\
\hline
LguidedLearn-1 &  84.01 & 97.72  \,  \, &  95.13   & 69.89  & 80.41  \\
LguidedLearn &  {\bf 85.58}  & {\bf 97.86 }  \,  \,  &  \bf{96.12 }  &  \bf{70.45 } &  \,  \bf{82.07 } \\
\hline
\end{tabular}
\label{tab:analysis-size-label-emb}
\end{table}

\paragraph{- Contextual and label attentive jointly learning analysis} 
LEAM model also performs label attentive learning (completed by a label and words jointly learning process) and is reported having effectiveness on the dataset of TCD-1 \cite{wang2018joint}. In our more experiments conducted on other dataset (such as TCD-2), the results illustrate that LEAM is not always effective and even much worse than other strong baselines (see Table \ref{tab:compare-TCD-2}). Besides publicated dataset, experimental results (details are not presented in the paper due to  space limitation) on our real application dataset also illustrate the problem.  The main reason is that the framework of LEAM is unqualified in encoding contextual information between words.  An important merit of our proposed LguideLearn framework is the contextual encoding and label attentive encoding jointly learning. In the framework, the label-guided layer (performing label attentive learning) can be easily and directly applied with an effective contextual learning model (BiLSTM) to achieve contextual information and label attentive constraints jointly encoding. Comparison experimental results are presented in Table \ref{tab:analysis-contextual-labelguided} to illustrate the importance of contextual and label attentive jointly learning. To control the effectiveness of using label embedding matrix, we make LguidedLearn also use one embedding for each label (as used in LEAM), denoted by {\bf LguidedLearn-1}. The results (in Table \ref{tab:analysis-contextual-labelguided}) show that even {\bf LguidedLearn-1} surpass LEAM by a large margin. At sometimes, the performance of LEAM is worse than a BiLSTM model (due to lacking of encoding contextual information between words), such as on the datasets of R8 and MR. 
%
%
\begin{table}[h!tb]
\centering
\caption{Compared to previous label embedding attentive learning model LEAM. LguidedLearn-1: LguidedLearn with using only one embedding to represent each label.}
	\begin{tabular}{@{}c@{}|c@{}c@{}c@{}c@{}c@{}}
    \hline
	{ Model } & \, {20NG}  \, &  {R8} \, &  {R52}  \, 	& {Ohsumed}  &	{MR}	 \\
\hline
BiLSTM &	\, 73.18  \,   \, & 96.31  \,  \,  &  90.54  \,  &  49.27  \, &  77.68 \\
LEAM &	81.91  & 93.31  & 91.84  & 58.58  & 76.95   \\  
LguidedLearn-1 &  {\bf 84.01} & {\bf 97.72}  &  {\bf 95.13}   & {\bf 69.89}  & {\bf 80.41}  \\
\hline
\end{tabular}
\label{tab:analysis-contextual-labelguided}
\end{table}

%
%
%



\subsection{Preliminary experiments with BERT}
Though we employ BiLSTM to encode contextual information in the framework LguidedLearn, our proposed label guided encoding layer actually can be applied with any other contextual information learner. Results of 
preliminary experiments with BERT are reported in Table \ref{tab:lguided-bert}.  In the experiments, we employ pre-trained BERT \footnote{\url{https://github.com/google-research/bert}} to produce contextual embeddings for each words in an input document/sentence. The very preliminary results presented in Table \ref{tab:lguided-bert} illustrate that the proposed label guided layer brings fruitful gains by performing label attentive learning.

\begin{table}[h!tb]
\centering
\caption{Label guided encoding layer is applied with BERT. Lguided-BERT-1: label guided encoding applied with the last layer of BERT. Lguided-BERT-3: label guided encoding applied with the last three layers of BERT, where each layer uses different label embedding matrix.}
	\begin{tabular}{@{}c@{}|c@{}c@{}c@{}c@{}c@{}}
    \hline
	{ Model } & \, {20NG}  \, &  {R8} \, &  {R52}  \, 	& {Ohsumed}  &	{MR}	 \\
\hline
BERT &	\, 67.90  \,   \, & 96.02  \,  \,  &  89.66  \,  &  51.17  \, &  79.24 \\
Lguided-BERT-1  &	 { 76.09}  & { 97.49}    &  { 94.26}    &  { 59.41}  &  { 81.03} \\ 
Lguided-BERT-3 &   {\bf 78.87} & {\bf 98.28}  &  {\bf 94.32}   & {\bf 62.37}  & {\bf 81.06}  \\
\hline
\end{tabular}
\label{tab:lguided-bert}
\end{table}

\section{Conclusion}
In this study, we propose a universal framework {LguidedLearn} to  exploit label global information for text representation and classification. In the framework, a label guided encoding layer can be easily and directly applied with a contextual information encoding module to bring fruitful gains  for text classification. A series of extensive experiments and analysis are presented to illustrate the effectiveness of our proposed learning schema. 

\clearpage
\clearpage

\bibliographystyle{named}
\bibliography{LguidedLearn-ijcai20}
\end{document}